\documentclass[10pt,twocolumn,letterpaper]{article}

\usepackage{iccv}
\usepackage{times}
\usepackage{epsfig}
\usepackage{graphicx}
\usepackage{amsmath}
\usepackage{amssymb}
\usepackage{graphicx}
\usepackage{amsmath}
\usepackage{amssymb}
\usepackage{booktabs}
\usepackage{times}
\usepackage{epsfig}
\usepackage{multirow}
\usepackage{amsfonts,amssymb}
\usepackage{subcaption}
\usepackage{algorithm}
\usepackage{algpseudocode}

\usepackage{amsmath,amsfonts,bm}

\def\eqref#1{equation~\ref{#1}}

\def\1{\bm{1}}

\DeclareMathAlphabet{\mathsfit}{\encodingdefault}{\sfdefault}{m}{sl}
\SetMathAlphabet{\mathsfit}{bold}{\encodingdefault}{\sfdefault}{bx}{n}

\DeclareMathOperator*{\argmin}{arg\,min}

\usepackage[pagebackref=true,breaklinks=true,letterpaper=true,colorlinks,bookmarks=false]{hyperref}

\iccvfinalcopy 

\def\iccvPaperID{5126} 

\ificcvfinal\fi

\begin{document}

\title{DiM: Distilling Dataset into Generative Model}

\author{
Kai Wang\textsuperscript{1}\thanks{Equal contribution.} 
\quad Jianyang Gu\textsuperscript{1,2}\footnotemark[1]
\quad Daquan Zhou\textsuperscript{1}
\quad Zheng Zhu\textsuperscript{3}
\quad Wei Jiang\textsuperscript{2}
\quad Yang You\textsuperscript{1}
\\
\textsuperscript{1}{National University of Singapore}
\quad \textsuperscript{2}{Zhejiang University}
\quad \textsuperscript{3}{Tsinghua University}
\\
\small{\texttt{\{gu\_jianyang, jiangwei\_zju\}@zju.edu.cn}}
\quad \small{\texttt{daquan.zhou@u.nus.edu}}\\
\small{\texttt{\{kai.wang, youy\}@comp.nus.edu.sg}}
\quad \small{\texttt{zhengzhu@ieee.org}}
\\
\small{Code: \url{https://github.com/vimar-gu/DiM}}
}

\maketitle
\ificcvfinal\thispagestyle{empty}\fi

\begin{abstract}
  Dataset distillation reduces the network training cost by synthesizing small and informative datasets from large-scale ones. Despite the success of the recent dataset distillation algorithms, three drawbacks still limit their wider application: i). the synthetic images perform poorly on large architectures; ii). they need to be re-optimized when the distillation ratio changes; iii). the limited diversity restricts the performance when the distillation ratio is large. In this paper, we propose a novel distillation scheme to \textbf{D}istill information of large train sets \textbf{i}nto generative \textbf{M}odels, named DiM. Specifically, DiM learns to use a generative model to store the information of the target dataset. During the distillation phase, we minimize the differences in logits predicted by a models pool between real and generated images. At the deployment stage, the generative model synthesizes various training samples from random noises on the fly.  Due to the simple yet effective designs, the trained DiM can be directly applied to different distillation ratios and large architectures without extra cost. We validate the proposed DiM across 4 datasets and achieve state-of-the-art results on all of them. To the best of our knowledge, we are the first to achieve higher accuracy on complex architectures than simple ones, such as 75.1\% with ResNet-18 and 72.6\% with ConvNet-3 on ten images per class of CIFAR-10. Besides, DiM outperforms previous methods with 10\% $\sim$ 22\% when images per class are 1 and 10 on the SVHN dataset.
\end{abstract}

\section{Introduction}
Deep learning models have shown superior performance in the computer vision area \cite{he2016deep, liu2021swin, dosovitskiy2020image, long2015fully, isola2017image, girshick2014rich, ren2015faster, he2017mask, he2015delving, simonyan2014very, simonyan2014two, parkhi2015deep}.
However, the success of these state-of-the-art models largely relies on ultra-large-scale datasets, which are hardly affordable for training with limited computational resources.
Dataset distillation (DD) \cite{wang2018dataset, zhao2021DC, wang2022cafe, zhaodm, zhao2021dsa}, as a promising approach, aims to reduce the heavy training cost through distilling discriminative information from large-scale datastes into small but informative ones. 

\begin{figure}[t]
\centering
\begin{subfigure}[]{\linewidth}
    \includegraphics[width = 1.0\linewidth]{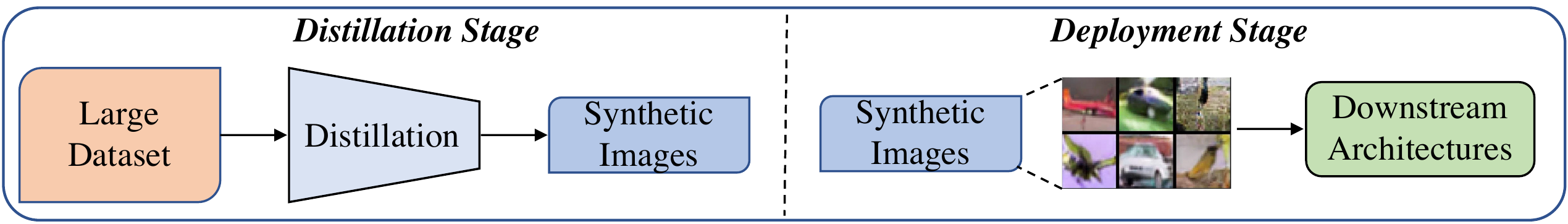}
    \caption{Paradigm of Distill knowledge into Images (DiI).}
    \label{fig:DiI_frame}
\end{subfigure}
\begin{subfigure}[]{\linewidth}
    \includegraphics[width = 1.0\linewidth]{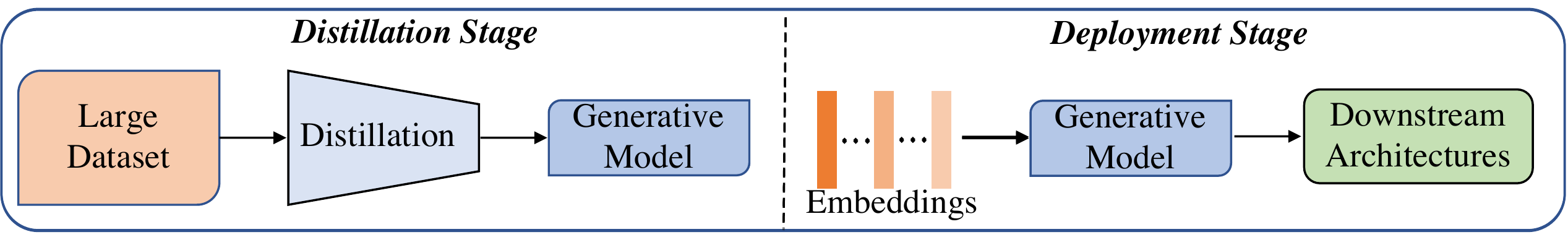}
    \caption{Paradigm of Distill knowledge into Model (DiM).}
    \label{fig:dim_frame}
\end{subfigure}
\begin{subfigure}{0.22\textwidth}
        \includegraphics[width=\linewidth]{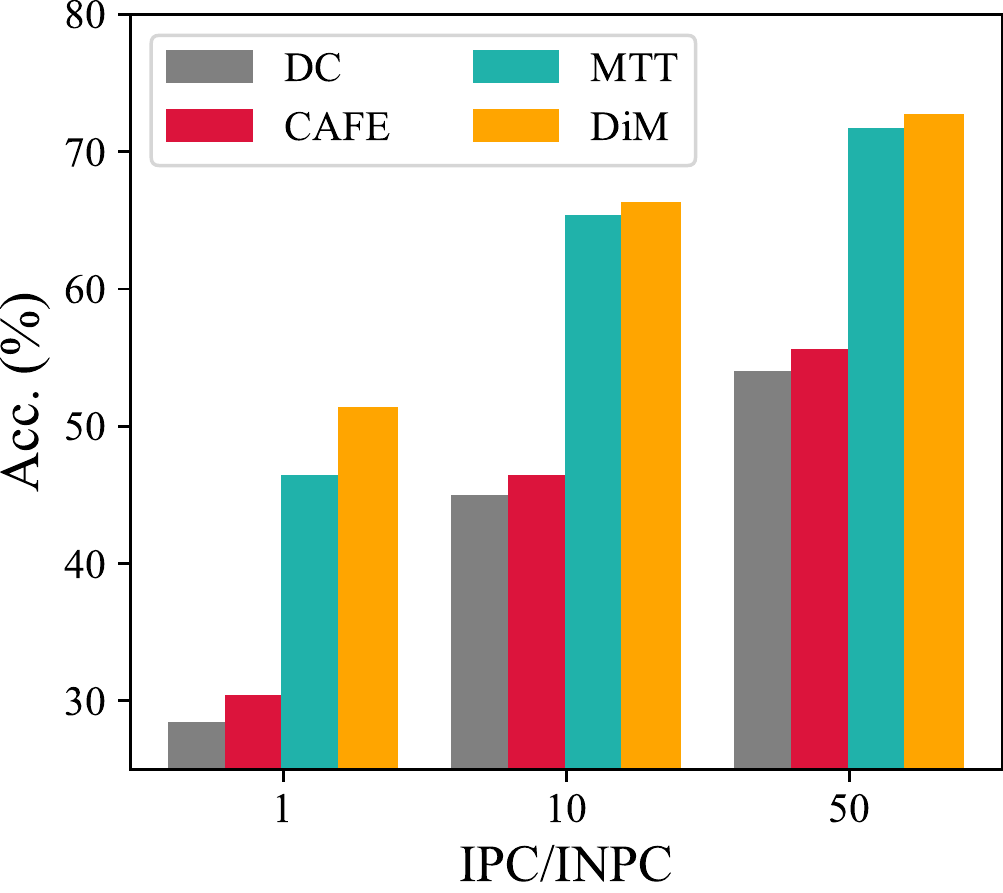}
        \caption{Performance comparisons.}
        \label{fig:acc_comparison}
    \end{subfigure}
    \begin{subfigure}{0.24\textwidth}
    \resizebox{\linewidth}{!}{
    \begin{tabular}{c|ccc}
        \toprule
        \multirow{2}{*}{Method}  & \multicolumn{2}{c}{Redeployment} \\
         &1 $\rightarrow$ 10 & 1 $\rightarrow$ 50  \\
         \hline
         DC~\cite{zhao2021DC} &1.5&7.6\\
         DSA~\cite{zhao2021dsa} &1.67&8.3 \\
         CAFE~\cite{wang2022cafe} &1.4&6.4\\
         DM~\cite{zhaodm} &1.3&6.1\\
         IDC~\cite{kimICML22} &18.2&87.1 \\
         MTT$^*$~\cite{cazenavette2022distillation} &8.3&40.9\\
         \textbf{DiM} &0.1&0.5\\
         \bottomrule
    \end{tabular}}
        \caption{GPU hours of redeployment.}
        \label{fig:cost_comparison}
    \end{subfigure}
\caption{(a) and (b): Illustrations of DiI and our DiM dataset distillation paradigms. $\mathcal{Z}$ and $\mathcal{Y}$ represent random noises and labels. (c): Comparisons of performances of DiI and DiM methods under different image number settings. (d): Comparisons of redeployment GPU hours of DiI and DiM methods. All results are obtained on CIFAR-10 with ConvNet-3. $^*$ denotes that the training time for the 2,000 checkpoints in MTT~\cite{cazenavette2022distillation} is not included. 
}

\label{fig:motivation}
\end{figure}

Current state-of-the-art dataset distillation methods \cite{wang2018dataset, zhao2021DC, wang2022cafe, cazenavette2022distillation, nguyen2021dataset, kimICML22} are based on an intuitive idea to match the status between synthetic and original sets.
There are mainly three types of matching strategies: gradient matching \cite{zhao2021DC, zhao2021dsa, kimICML22}, training trajectory matching \cite{cazenavette2022distillation}, and feature or distribution matching \cite{zhaodm, wang2022cafe}.
We note the above methods as DiI since they aim to \textbf{D}istill knowledge \textbf{i}nto \textbf{I}mages (as shown in Fig. \ref{fig:DiI_frame}).

Although these DiI strategies achieve remarkable performances in the dataset distillation area, they have the following shortcomings: 1). Most previous DiI works~\cite{zhao2021DC, zhao2021dsa, zhaodm, cazenavette2022distillation, kimICML22, yang2022dery} conduct distillation based on 3-layer convolutional networks. The synthesized dataset performs poorly on unseen architectures, especially larger ones, such as ResNet~\cite{he2016deep}, VGG~\cite{simonyan2014very}, and DenseNet~\cite{huang2017densely}.
2). DiI methods store information in synthetic images. When the Image Per Class (IPC) setting or distillation ratio changes, DiI methods usually need to retrain the distillation stage.
We compare performances and redeployment costs of several DiI methods in Fig. \ref{fig:acc_comparison} and \ref{fig:cost_comparison}. 
One can find DiI methods perform poorly on small IPC settings.
Besides, redeploying IPC = 1 to 10 costs about 8.3 hours for MTT~\cite{cazenavette2022distillation} and 18.2 hours for IDC~\cite{kimICML22}, which is a heavy burden.

As illustrated in Fig. \ref{fig:dim_frame}, in this paper, we aim to \textbf{D}istill information of original datasets \textbf{i}nto a \textbf{M}odel (DiM) instead of images.
Specifically, given labels, we utilize a conditional generative model~\cite{mirza2014conditional} to reconstruct training images from latent space and simultaneously distill knowledge from original datasets. 
To make the generated images helpful to the classification task, we propose a logits-based matching strategy to minimize the differences of prediction logits between real and generated images.
Considering various supervisions are crucial for discriminative feature learning~\cite{zhou2022dataset}, the logits is extracted by a model that is randomly selected from a models pool in each epoch.  
At the deployment stage, the generative model synthesizes various training samples from random noises on the fly. 
We summarize the differences between DiI and DiM in Tab.~\ref{tab:diff_bt_dii_dim}. One can find that DiM has obvious advantages in cross-architecture generalization and redeployment efficiency. 
As DiM stores information in models rather than images, directly using IPC as a budget leads to insufficient utilization. 
To sufficiently demonstrate the learned knowledge of DiM in the distillation stage, we define Image Numbers of Per Class (INPC) to keep the same forward number as DiI for comparison.

\begin{table}[t]
    \centering
    \caption{The comparisons between DiI and DiM.}
    \label{tab:diff_bt_dii_dim}
    \small
    \begin{tabular}{c|c|c}
    \toprule
         Comparison Item& DiI & DiM \\
         \midrule
         Information Container & Images & Models \\
         Cross-arch. Performance & Unstable & Stable \\
         Cross-IPC/INPC Redeployment & Every time & Once for all \\
         \bottomrule
         
    \end{tabular}
\end{table}

DiM can be easily redeployed with different distillation ratios and architectures thanks to simple yet efficient designs.
We conduct experiments on several popular benchmarks and demonstrate that the results yielded by DiM have the following benefits.
First, the results yielded by DiM are significantly superior to state-of-the-art methods in all settings, especially on large architectures.
Second, as illustrated in Fig. \ref{fig:cost_comparison}, DiM saves 13$\times$ $\sim$ 160$\times$ redeployment cost when compared with DiI methods.
Third, with rich information from models pool, DiM shows more robust cross-architecture generalization than DiI methods.

The main contributions of this paper are summarised as:

\begin{itemize}
    \item We propose DiM, a novel scheme that distills the dataset into a generative model instead of images, with significant improvements over previous DiI methods in large architecture scalability, redeployment efficiency, and cross-architecture generality.
    
    \item In DiM, models pool and logits matching are proposed to distill knowledge from the original dataset, where models pool provides rich supervisions and logits matching aims to minimize the logits differences of real and synthetic images. 
    
    \item DiM consistently outperforms state-of-the-art DiI methods on various datasets and architectures with 13$\times$ $\sim$ 160$\times$ lower redeployment cost, showing its effectiveness and generality for dataset distillation. 
\end{itemize}

\section{Related Works}
\subsection{Dataset Distillation.}
Wang \textit{et.al} is the first to propose the definition of Dataset Distillation (DD) \cite{wang2018dataset}: synthesize a small-scale set from a large one and achieve comparable training results.
Specifically, they propose a model to generate synthetic samples that are subject to minimization of the training loss differences between synthetic and original data.
After that, 
DC \cite{zhao2021DC}, DSA \cite{zhao2021dsa}, and IDC \cite{kimICML22} are proposed to match gradients between synthetic and original samples. 
CAFE \cite{wang2022cafe}, and DM \cite{zhaodm} introduce a feature matching paradigm to reduce the influence of synthetic images that bias to large-gradient samples from the original set.
MTT \cite{cazenavette2022distillation} proposes to minimize the loss of training trajectories between synthetic and original images.
HaBa \cite{yang2022dery} divides the original dataset into bases and several hallucinators to keep the diversity of synthetic images.
FRePo ~\cite{zhou2022dataset} proposes neural feature regression with pooling to reduce the computation and memory costs of DD.
MAP \cite{deng2022remember} designs a method where the information of the original dataset is stored in bases and addressing matrices.
Although these methods achieve state-of-the-art results in the dataset distillation area, they are still faced with the cross-architecture generalization and redeployment cost issues. Our work is concurrent with \cite{zhang2023dataset}.

\begin{figure*}[htp]
    \centering
    \includegraphics[width=1.0\textwidth]{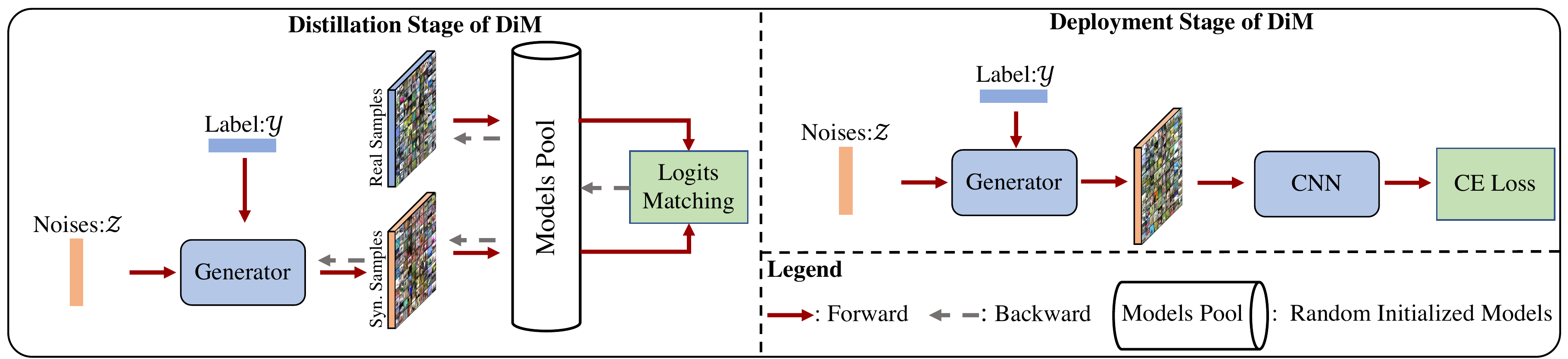}
    \caption{Illustration of proposed `Distill into Model' (DiM) framework. DiM consists of two stages the same as previous DiI methods, named distillation and deployment stages. Distillation stage has two special designs: models pool and logits matching (LM). Models pool includes several randomly initialized models on target dataset from open-source community. LM is regarded as bridge to distill knowledge from target dataset to generator. $\mathcal{Z}$ and $\mathcal{Y}$ denote the random noises and labels.}
    \label{fig:pipeline}
\end{figure*}

\subsection{Generative Models}
Our work is also related to generative adversarial networks (GANs) \cite{goodfellow2020generative, xia2022gan, karras2020analyzing, shen2020interpreting, goodfellow2014generative}.
\cite{goodfellow2020generative} is the first proposal to generate real-looking images.
After that, many GAN-based methods have been proposed for other tasks, such as cycleGAN ~\cite{isola2017image} introduces a bi-directional GANs for style transform, InfoGAN ~\cite{chen2016infogan} maximizes the mutual information between a small subset of the latent
variables and the observation, and BigGAN ~\cite{brock2018large} studies training GANs on largest scale datasets.
The main difference is that our generation model aims to distill discriminate information from the original dataset while GANs merely target to synthesize real-looking images.
GAN Inversion \cite{xia2022gan, karras2020analyzing, shen2020interpreting} aims to find the inputs of latent space for the pre-trained GAN model, which can faithfully recover a given image. 
ITGAN \cite{zhao2022synthesizing} optimizes a fixed number of inputs for a pre-trained generator, but it needs to retrain when INPC changes.

\section{Method}
In this section, we first review preliminaries of dataset distillation \cite{wang2018dataset, zhao2021DC, zhao2021dsa, wang2022cafe, kimICML22, deng2022remember} and GANs \cite{goodfellow2020generative, mirza2014conditional}.
Then we introduce DiM and its components.
Finally, we discuss the differences between GANs, GAN Inversion, and DiM.

\subsection{Preliminaries}
We briefly introduce preliminaries of dataset distillation (DD) and generative adversarial networks (GANs).

\textbf{DD.}
DD aims to synthesize a small yet informative dataset that achieves comparable training results to the original large dataset.
Given a large dataset $\mathcal{T} = {(x_i, y_i)}|_{i=1}^{|\mathcal{T}|}$, DD designs algorithms to distill the information into a small dataset $\mathcal{S} = {(s_j, y_j)}|_{j=1}^{|\mathcal{S}|}$ ($|\mathcal{S}| \ll |\mathcal{T}|$). The distillation operation can be formulated as follows,

\begin{equation}
    \mathcal{S} = f(\mathcal{T}; \mathcal{S}_{0}; \theta),
\end{equation}
where $f$ denotes the distillation algorithm with parameter $\theta$, $\mathcal{S}_{0}$ represents the initialization of synthetic data, and $\mathcal{S}$ denotes the final synthetic image dataset. 
The knowledge of the original train set is distilled into $\mathcal{S}$ during the distillation process.
After the distillation stage, $\mathcal{S}$ is deployed on downstream tasks, such as classification, to evaluate its performance, denoted as the deployment stage. 

\textbf{GANs.}
The goal of GANs \cite{goodfellow2020generative} is to synthesize photo-realistic images, which includes a generator $\mathcal{G}$ and discriminator $\mathcal{D}$.
In GANs training, the generator and discriminator are jointly optimized for the mini-max loss function as,
\begin{equation}
    \min_\mathcal{G} 
    \max_\mathcal{D} 
    \mathbb{E}_{\mathcal{X}\sim P(\mathcal{X})}[\log \mathcal{D}(\mathcal{X})] + \mathbb{E}_{\mathcal{Z}\sim P(\mathcal{Z})}[\log (1- \mathcal{D}(\mathcal{G}(\mathcal{Z})))],
\end{equation}
where $P(\mathcal{Z})$ and $P(\mathcal{X})$ denote the distribution of latent vector $\mathcal{Z}$ and real images.
Conditional GANs~\cite{mirza2014conditional} are proposed to introduce the labels $\mathcal{Y}$ into the generation process for more specific generation targets.
The optimization loss of conditional GANs can be written as,

\begin{equation}
\begin{split}
     \min_\mathcal{G} 
    \max_\mathcal{D} 
    \mathbb{E}_{\mathcal{X}\sim P(\mathcal{X})}[\log \mathcal{D}(\mathcal{X}|\mathcal{Y})] \\ + \mathbb{E}_{\mathcal{Z}\sim P(\mathcal{Z})}[\log (1- \mathcal{D}(\mathcal{G}(\mathcal{Z}|\mathcal{Y})))].
\end{split}
\label{eq:cgan}
\end{equation}
In this paper, we utilize conditional GANs~\cite{mirza2014conditional} for generating images. However, our motivation is different from current conditional GAN methods. We aim to encourage the generator to synthesize images that are helpful for downstream classification training.

\subsection{Overview of DiM}
We propose a novel scheme, `Distill into Models' (DiM), for the dataset distillation task.
The most significant difference between DiI and DiM is the information container (Images or Model).
The pipeline of our proposed DiM is illustrated in Fig. \ref{fig:pipeline}.
DiM also consists of the same two stages as DiI: distillation and deployment.
The distillation stage aims to train a generator that can synthesize discriminative images from random noises and labels.
The deployment stage is designed for evaluating the trained generator.
For each batch, we randomly sample $B$ paired random noises $\mathcal{Z}$ and labels $\mathcal{Y}$.
Then, the paired $\mathcal{Z}$ and $\mathcal{Y}$ are fed into the generator to generate images.
After that, we sample $B$ real images with the same labels of $\mathcal{Y}$ from the original dataset.
A model is randomly sampled from models pool to predict the classification logits of real and synthetic images, respectively.
Minimizing the logits differences is served as the optimization target for distilling the knowledge from real images into the generative model.
At the deployment stage, the generative model synthesizes various training samples from random noises on the fly. 

\subsection{Distill into Model}
\textbf{Synthetic Images Generation.}
Given a batch of random noises $\mathcal{Z} \in \mathbb{R}^{B \times K}$ and corresponding labels $\mathcal{Y} \in \mathbb{R}^{B \times C}$, a generator model is utilized to synthesize images. This process can be formulated as,

\begin{equation}
    \mathcal{S}_{\mathrm{DiM}} = \mathcal{G}([\mathcal{Z} \oplus \mathcal{Y}]; \beta),
\end{equation}
where $\oplus$ represents the concatenation operation, $\beta$ is the parameters of $\mathcal{G}$, $B$ is the batch size, and $C$ denotes the number of categories. $\mathcal{Y}$ is transformed into one-hot embedding for convenience. $K$ is the dimension of random noise, and we evaluate its impact in supplementary materials.

\textbf{Logits Matching.}
Different from previous works ~\cite{zhao2021DC, zhao2021dsa, cazenavette2022distillation, wang2022cafe}, we utilize a Logits Matching (LM) strategy instead of matching the gradients, features/distributions, or training trajectories.
The modification is based on the following considerations: 1). These three matching strategies perform well only when optimized category by category, which reduces the efficiency of DD; 2). Gradient matching is easily biased to samples with large gradients in the original train set~\cite{wang2022cafe}. 3). Feature/distribution matching is designed to minimize the distance of mean features/distribution in each batch, where the diversity of synthetic images is not constrained; 4). Training trajectory matching is time-consuming for pre-training a large number of models (\textit{i.e.} 2, 000 checkpoints).
Hence, we match the prediction logits for real and synthetic images.
For a batch of real images $I_{r}$ and synthetic images  $\mathcal{S}_{\mathrm{DiM}}$, we apply a model $m$ to predict the classification logits of these images.
Formally, logits matching can be written as follows,

\begin{equation}
    L_{m} = \mathrm{MSE}(m(\mathcal{S}_{\mathrm{DiM}}); m(I_{r})),
\end{equation}
where $\mathrm{MSE}(\cdot; \cdot)$ denotes the mean-square error loss. 
LM aims to minimize the prediction logits that influences the results of downstream classification task directly.
Therefore, LM provides more specific supervisions when compared to previous matching strategies. 
Besides, logits matching does not need to be optimized category by category, which largely improves training efficiency in the distillation stage.

\textbf{Models Pool.}
Previous works \cite{wang2022cafe, zhao2021DC, zhao2021dsa} only use one model to match the difference between real and synthetic images, which provides limited supervision for informative image synthesis.
MTT \cite{cazenavette2022distillation} trains 2,000 checkpoints as a trajectory buffer in advance, which results in heavy computational and memory cost.
Here, we argue that supervisions diversified in the architecture dimension are beneficial for distilling the information from original train dataset.
Therefore, we extend the matching model to a models pool.
In each epoch, we randomly select one random-initialized model from the models pool. The selection operation is formulated as follows,
\begin{equation}
    m = \mathcal{M}[\mathrm{Random.choice}(|\mathcal{M}|, 1)],
\end{equation}
where $\mathcal{M}$ denotes the models pool, $m$ is the selected model.
Our models pool has two benefits:
1). Models pool provides various views for logits matching, which helps to distill more discriminative information from original dataset.
2). Models pool can also mitigate the over-fitting on one architecture~\cite{zhou2022dataset}, leading to superior generalization performances over previous works~\cite{zhao2021DC, zhaodm, zhao2021dsa, wang2022cafe, cazenavette2022distillation, kimICML22} largely.

\begin{table*}[tp]
    \caption{The performance (\%) comparison to state-of-the-art methods. LD$^{\dag}$ and DD$^{\dag}$ use AlexNet for CIFAR10, while the rest use ConvNet for training and testing. Underline denotes results from \cite{yang2022dery}. \textbf{Bold entries} are best results.}
\label{tab:compare_sota}
    \centering
    \scriptsize
    \setlength{\tabcolsep}{3pt}
    \begin{tabular}{cc|ccc|ccc|ccc|ccc}
    \toprule
    \multirow{3}{*}{} & Dataset & \multicolumn{3}{c|}{SVHN} & \multicolumn{3}{c|}{CIFAR10} & \multicolumn{3}{c|}{MNIST} & \multicolumn{3}{c}{FashionMNIST} \\
    \midrule
    & IPC & 1 & 10 & 50 & 1 & 10 & 50 & 1 & 10 & 50 & 1 & 10 & 50 \\
    & Ratio \% & 0.014 & 0.14 & 0.7 & 0.02 & 0.2 & 1 & 0.2 & 2 & 10 & 0.2 & 2 & 10 \\ \midrule
    \multirow{4}{*}{Coreset} & Random & 14.6$\pm$1.6 & 35.1$\pm$4.1 & 70.9$\pm$0.9 & 14.4$\pm$2.0 & 26.0$\pm$1.2 & 43.4$\pm$1.0 & 64.9$\pm$3.5 & 95.1$\pm$0.9 & 97.9$\pm$0.2 &51.4$\pm$3.8  & 73.8$\pm$0.7 & 82.5$\pm$0.7  \\
    & Herding & 20.9$\pm$1.3 & 50.5$\pm$3.3 & 72.6$\pm$0.8 & 21.5$\pm$1.3 & 31.6$\pm$0.7 & 40.4$\pm$0.6 & 89.2$\pm$1.6 & 93.7$\pm$0.3 & 94.8$\pm$0.2  &67.0$\pm$1.9  & 71.1$\pm$0.7 & 71.9$\pm$0.8 \\
    & K-Center & 21.0$\pm$1.5 & 14.0$\pm$1.3 & 20.1$\pm$1.4 & 21.5$\pm$1.3 & 14.7$\pm$0.9 & 27.0$\pm$1.4 & 89.3$\pm$1.5 & 84.4$\pm$1.7 & 97.4$\pm$0.3 &66.9$\pm$1.8  & 54.7$\pm$1.5 & 68.3$\pm$0.8 \\
    & Forgetting & 12.1$\pm$1.7 & 16.8$\pm$1.2 & 27.2$\pm$1.5 & 13.5$\pm$1.2 & 23.3$\pm$1.0 & 23.3$\pm$1.1 & 35.5$\pm$5.6 & 68.1$\pm$3.3 &  88.2$\pm$1.2 &42.0$\pm$5.5  & 53.9$\pm$2.0 & 55.0$\pm$1.1  \\
    \midrule
    \multirow{10}{*}{DiI} & DD$^{\dag}$~\cite{wang2018dataset} & - & - & - & - & 36.8$\pm$1.2 & - & - & 79.5$\pm$8.1 & -  \\
    & LD$^{\dag}$~\cite{bohdal2020flexible} & - & - & - & 25.7$\pm$0.7 & 38.3$\pm$0.4 & 42.5$\pm$0.4 & 60.9$\pm$3.2 &  87.3$\pm$0.7 &  93.3$\pm$0.3  \\
    & DC~\cite{zhao2021DC} & 31.2$\pm$1.4 & 76.1$\pm$0.6 & 82.3$\pm$0.3 & 28.3$\pm$0.5 & 44.9$\pm$0.5 & 53.9$\pm$0.5 & 91.7$\pm$0.5 & 97.4$\pm$0.2 & 98.8$\pm$0.2  &70.5$\pm$0.6  & 82.3$\pm$0.4 & 83.6$\pm$0.4  \\
    & DSA~\cite{zhao2021dsa} & 27.5$\pm$1.4 & 79.2$\pm$0.5 & 84.4$\pm$0.4 & 28.8$\pm$0.7 & 52.1$\pm$0.5 & 60.6$\pm$0.5 & 88.7$\pm$0.6 & 97.8$\pm$0.1 & 99.2$\pm$0.1  &70.6$\pm$0.6  & 84.6$\pm$0.3 & 88.7$\pm$0.2 \\
    & DM~\cite{zhaodm} & - & - & - & 26.0$\pm$0.8 & 48.9$\pm$0.6 & 63.0$\pm$0.4 & 89.7$\pm$0.6 & 97.5$\pm$0.1 & 98.6$\pm$0.1 &- &- &-  \\
    & CAFE~\cite{wang2022cafe} & 42.6$\pm$3.3 & 75.9$\pm$0.6 & 81.3$\pm$0.3 & 30.3$\pm$1.1 & 46.3$\pm$0.6 & 55.5$\pm$0.6 & 93.1$\pm$0.3 & 97.2$\pm$0.2 & 98.6$\pm$0.2 & 77.1$\pm$0.9 & 83.0$\pm$0.4 & 84.8$\pm$0.4   \\
    & CAFE+DSA~\cite{wang2022cafe} & 42.9$\pm$3.0 & 77.9$\pm$0.6 & 82.3$\pm$0.4 & 31.6$\pm$0.8 & 50.9$\pm$0.5 & 62.3$\pm$0.4 & 90.8$\pm$0.5 & 97.5$\pm$0.1 & 98.9$\pm$0.2 & 73.7$\pm$0.7 & 83.0$\pm$0.3 & 88.2$\pm$0.3   \\
     & KIP~\cite{nguyen2021dataset} & 57.3$\pm$0.1 & 75.0$\pm$0.1 & 80.5$\pm$0.1 & 49.9$\pm$0.2 & 62.7$\pm$0.3 & 68.6$\pm$0.2 & 90.1$\pm$0.1 & 97.5$\pm$0.0 & 98.3$\pm$0.1 & 73.5$\pm$0.5 & 86.8$\pm$0.1 & 88.0$\pm$0.1 \\
    & MTT~\cite{cazenavette2022distillation} & \underline{58.5$\pm$1.4} & \underline{70.8$\pm$1.8} & \underline{85.7$\pm$0.1} & 46.3$\pm$0.8 & 65.3$\pm$0.7 & 71.6$\pm$0.2 & - & - & - &- &- &-  \\
     & FRePo~\cite{zhou2022dataset} & - & - & - & 46.8$\pm$0.7 & 65.5$\pm$0.4 & 71.7$\pm$0.2 & 93.0$\pm$0.4 & 98.6$\pm$0.1 & 99.2$\pm$0.0 &75.6$\pm$0.3 &86.2$\pm$0.2 &89.6$\pm$0.1  \\
    \midrule
    & INPC & 1 & 10 & 50 & 1 & 10 & 50 & 1 & 10 & 50 & 1 & 10 & 50 \\
    \midrule
    \multirow{1}{*}{GAN} & Conditional GAN~\cite{mirza2014conditional} & 78.3$\pm$0.9 & 83.8$\pm$0.6 & 86.6$\pm$0.5 & 46.4$\pm$1.2 & 62.7$\pm$0.9 & 68.1$\pm$0.8 & 96.1$\pm$0.7 & 97.8$\pm$0.3 & 98.4$\pm$0.3 & 81.5$\pm$0.5 & 84.0$\pm$0.2 & 86.3$\pm$0.3 \\
    \midrule
    DiM & DiM & \textbf{80.9$\pm$1.2} & \textbf{87.9$\pm$0.5} & \textbf{90.4$\pm$0.3} & \textbf{51.3$\pm$1.0} & \textbf{66.2$\pm$0.5} & \textbf{72.6$\pm$0.4} & \textbf{96.5$\pm$0.6} & \textbf{98.6$\pm$0.2} & \textbf{99.2$\pm$0.2} & \textbf{84.5$\pm$0.4} & \textbf{88.2$\pm$0.2} & \textbf{89.8$\pm$0.1} \\
    \midrule
    \multicolumn{2}{c|}{Whole Dataset} & \multicolumn{3}{c|}{95.4$\pm$0.1} & \multicolumn{3}{c|}{84.8$\pm$0.1} & \multicolumn{3}{c|}{99.6$\pm$0.0} & \multicolumn{3}{c}{93.5$\pm$0.1}  \\
    \bottomrule
    \end{tabular}
\end{table*}

\begin{algorithm}[t]
\caption{Optimization of DiM. }
\label{alg-opt}
\begin{algorithmic}
\State \textbf{Input:} 
$N$ denotes the number of training epoch for vanilla GANs. $J$ is the total iteration numbers for each epoch. $\epsilon$ is learning rates. The optimizer is SGD.
\State For $i = 1, \ldots, N$: \Comment{Train generator with GAN loss.}
\State \qquad  For $j = 1, \ldots, J$: \Comment{Iterations of each epoch.}
\State \qquad \qquad Calculating $L_{g}$.
\State \qquad \qquad $\beta$ $\leftarrow$ $\beta - \epsilon\frac{\partial L_{g}}{\partial \beta}$.  \Comment{Update generator.}
\State For $q = 1, \ldots, Q$  \Comment{Add $L_{m}$ for $Q$ epochs.}
\State \qquad  For $j = 1, \ldots, J$: \Comment{Iterations of each epoch.}
\State \qquad \qquad Calculating $L_{g} + \lambda L_{m}$.
\State \qquad \qquad $\beta$ $\leftarrow$ $\beta - \epsilon\frac{\partial (L_{g} + \lambda L_{m})}{\partial \beta}$.  \Comment{Update generator.}
\State \textbf{Output:} Saving the parameters $\beta$ of generator. 
\end{algorithmic}
\end{algorithm}

\textbf{Multi-task Optimization.}
As a randomly initialized generator produces almost noise outputs, which is unable to form valid matching with real images, we first train vanilla GANs for $N$ epochs. In this stage, only GAN loss is used to optimize the network. Then, we jointly use $L_{m}$ and GAN loss to optimize the networks. The total loss is written as, 
\begin{equation}
    L_{total} = L_{g} + \lambda L_{m},
\end{equation}
where $L_{g}$ represents the GAN loss and $\lambda$ is the trade-off parameter between $L_{g}$ and $L_{m}$.
We study the influence of $\lambda$ in the experiment section. After $N$th epoch, the parameters of generator $\beta$ is updated by minimizing $L_{total}$:
\begin{equation}
    \beta \leftarrow \argmin_\beta L_{total}.
\end{equation}

To better understand our DiM, we summarize the training algorithm in Alg. \ref{alg-opt}
\subsection{Discussions with GANs and GAN Inversion}
Our DiM is different from GANs \cite{goodfellow2020generative, goodfellow2014generative} in the optimization targets.
DiM aims to optimize a generator that can synthesize images usable for downstream tasks, while vanilla GANs only generates visually real images.
GAN Inversion methods \cite{xia2022gan, karras2020analyzing, shen2020interpreting} are designed to find the $\mathcal{Z}$ in the latent space with original real images for a pre-trained GAN model.
During the inversion stage, $\mathcal{Z}$ is optimized by the following equation,
\begin{equation}
    \mathcal{Z}^* \leftarrow \argmin_\mathcal{Z} ||\mathcal{G}(\mathcal{Z}) - \mathcal{I}||,
\end{equation}
where $\mathcal{I}$ denotes the images from training dataset and $\mathcal{G}$ is the generator.
Note that the parameters of $\mathcal{G}$ are fixed in inversion training.
Our DiM is different from GAN Inversion-based methods in the following aspects: First, the goals of these two methods are different. DiM aims to optimize a generator to generate discriminate images for training, while GAN Inversion is designed to learn a reverse mapping between real images and the latent embeddings.
Second, the initialization for the generator is different. GAN Inversion optimizes latent embeddings while fixes the trained generator. Our DiM optimizes a randomly initialized generator and expects the generator to synthesize discriminate images from random noises.
Third, the stored information is different. DiM only stores the generative model. Oppositely, GAN Inversion is required to store both the latent embeddings and the generative model.

\section{Experiments}
\label{exp}

In this section, we conduct extensive experiments with DiM on a variety of datasets and architectures to demonstrate the effectiveness and generality of DiM.

\subsection{Datasets and Implementation Details}
Following previous works, we evaluate the effectiveness of DiM on MNIST~\cite{lecun1998gradient}, FashionMNIST~\cite{xiao2017fashion}, SVHN~\cite{sermanet2012convolutional}, and CIFAR-10~\cite{krizhevsky2009learning}.
MNIST is a digital handwritten number (0 to 9) dataset. It contains 60,000 training samples and 10,000 testing samples with 28 $\times$ 28 resolution. FashionMNIST consists of 60,000 training images and 10,000 testing images. All images are with 28 $\times$ 28 resolution and are from 10 categories.
SVHN is captured from street view, containing 73,257 training digits and 26,032 testing ones from 10 classes. 
CIFAR-10 contains colored natural images with the size of 32 $\times$ 32 of 10 categories. There are 50,000 images for training and 10,000 images for testing.

\textbf{Implementation Details.}
We here present the details of distillation and deployment stages, respectively. In distillation stage, we first train a conditional GANs for $N$ epochs. We defaultly set $ N = 120$. For hyper parameters of conditional GANs, we employ Adam~\cite{kingma2015adam} with a learning rate $\epsilon$ = 1e-4. The batch size $B$ is 64. The trade-off $\lambda$ is set to 0.01. The sensitiveness of $N$ and $\lambda$ are investigated in Sec. \ref{sec:lambda}.
In the deployment stage, we save the trained generator $\mathcal{G}$. Random noises and labels are sampled as inputs for $\mathcal{G}$. Following the setting in IDC~\cite{kimICML22}, we train the validation model for 1,000 epochs.
The optimizer is set as SGD, with a learning rate of 0.01.
For data augmentation, we adopt the settings in DSA~\cite{zhao2021dsa}.
As our DiM aims to distill knowledge from original datasets and memorize it into a model, we conduct the same forward numbers for fair comparison. 

\subsection{Comparison to coreset and DiI Methods}
As shown in Tab. \ref{tab:compare_sota}, we compare our DiM with coreset selection, DiI methods, and vanilla GAN.
It should be noted that INPC is not proposed for unfair comparison but for sufficiently demonstrating the knowledge learned from vanilla GAN and DiM.
We list the results of state-of-the-art coreset and DiI methods. 
Based on previous works' results in Tab. \ref{tab:compare_sota}, we have following observations:
1). Coreset or randomly selection perform poorly as the information diversity of corset or subset is poor in the dataset distillation setting (\textit{i.e.} $|\mathcal{S}| \ll |\mathcal{T}|$). 
2). The performances of DiI-based methods largely rely on the number of IPC. The performance drops significantly when IPC is small, such as IPC = 1.
Our proposed DiM achieves state-of-the-art performances on all datasets and INPCs.
Especially when INPC is equal to 1, the improvement of DiM is up to 22\%, which demonstrates the strong information capacity of the generator.
\begin{table}[t]
\renewcommand\arraystretch{1.0}
\caption{Test performance comparisons among HaBa ~\cite{yang2022dery}, IDC ~\cite{kimICML22}, and DiM. To make fair comparisons, we keep the same INPC$^{*}$ in each epoch. \textbf{Bold entries} are best results.}
\centering
\small
\setlength{\tabcolsep}{5pt}

\begin{tabular}{cc|ccc|c}
\toprule
\multirow{2}{*}{}           & \multirow{2}{*}{INPC$^{*}$} 
&\multicolumn{3}{c|}{Condensation}    & Whole \\ 
                            & & HaBa &IDC & DiM  & Dataset     \\ \midrule
\multirow{3}{*}{MNIST}          & 1   &-&94.2&\textbf{97.3} & \multirow{3}{*}{99.6} \\ 
                                & 10    &-&98.4&\textbf{98.7}   &\\ 
                                & 50    &-&99.1&\textbf{99.2}  & \\ \midrule

\multirow{3}{*}{FashionMNIST}   & 1     &-&81.0&\textbf{86.0}  & \multirow{3}{*}{93.5} \\ 
                                & 10    &-&86.0&\textbf{89.0}   &         \\ 
                                & 50     &-&86.2&\textbf{90.2} & \\ \midrule

\multirow{3}{*}{SVHN}           & 1     &69.8 &68.5 &\textbf{83.8}  & \multirow{3}{*}{95.4} \\ 
                                & 10   &83.2&87.5 &\textbf{89.1}  &     \\
                                & 50   &88.3&90.1 &\textbf{91.0}  & \\ \midrule 

\multirow{3}{*}{CIFAR10}        & 1      &48.3&50.6 &\textbf{55.8}  & \multirow{3}{*}{84.8}         \\ 
                                & 10   &69.9&67.5 &\textbf{70.8}  &         \\ 
                                & 50     &74.0&74.5 &\textbf{74.9} & \\ 
                            \bottomrule
\end{tabular}

\label{tab:sota_with_idc}
\end{table}

\textbf{Comparisons to HaBa and IDC.}
HaBa ~\cite{yang2022dery} divides the original dataset into bases and transformation functions (\textit{i.e.} style transform network). For example, HaBa ~\cite{yang2022dery} takes consideration of memory as a budget and only saves single-channel synthetic images.
IDC ~\cite{kimICML22} defaults to set factor as 2 for training, so a single image consists of two sub-images. 
Therefore, the forward number of these methods is different.
To make a fair comparison to them, we increase the forward number to the same. The comparison results are shown in Tab. \ref{tab:sota_with_idc}. One can find that DiM achieves the highest results on all datasets, which shows the strong effectiveness of our proposed method.

\subsection{Ablation Studies}
In this section, we perform extensive ablation studies to illustrate the effects of our method. For better evaluation, we conduct experiments on CIFAR-10 with INPC = 50 if not otherwise stated.

\begin{figure*}[htp]
    \centering
    \begin{subfigure}{0.33\textwidth}
      {\includegraphics[width=0.95\linewidth,height=0.82\linewidth]{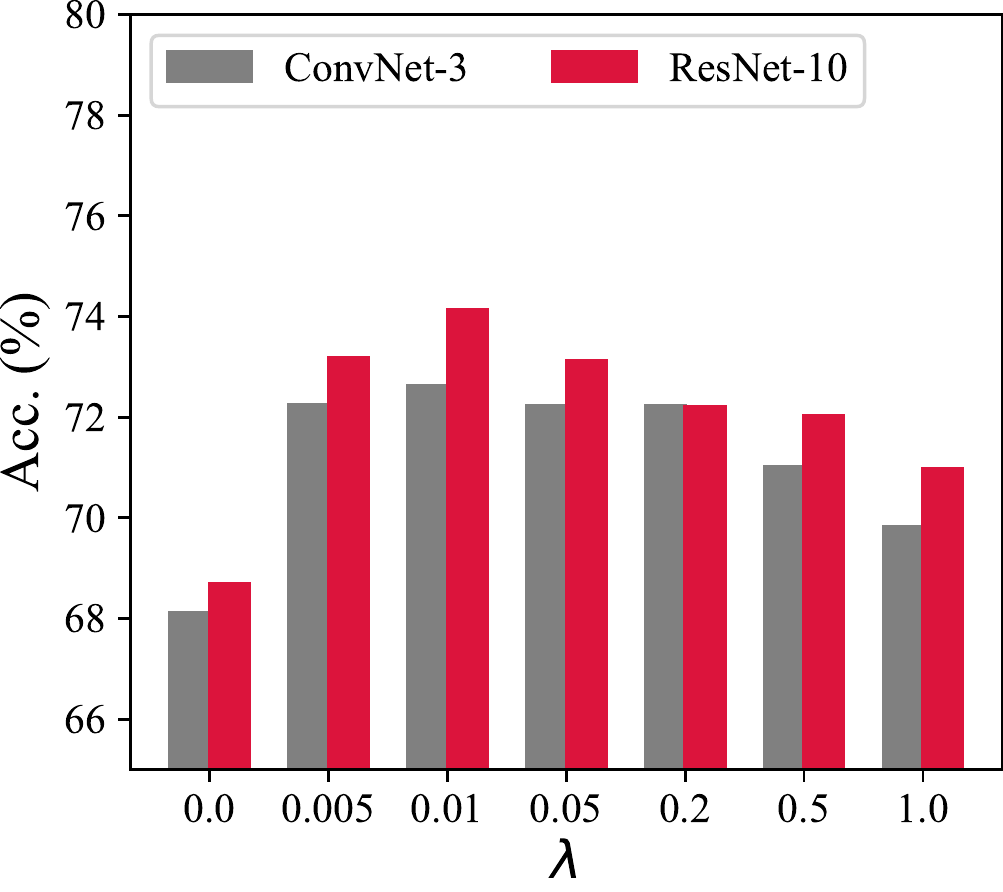}}
      
      \caption{Evaluation of $\lambda$.}
      \label{fig:sensitive_of_lbd}
    \end{subfigure}
\begin{subfigure}{0.33\textwidth}
    \includegraphics[width=0.95\linewidth,height=0.82\linewidth]{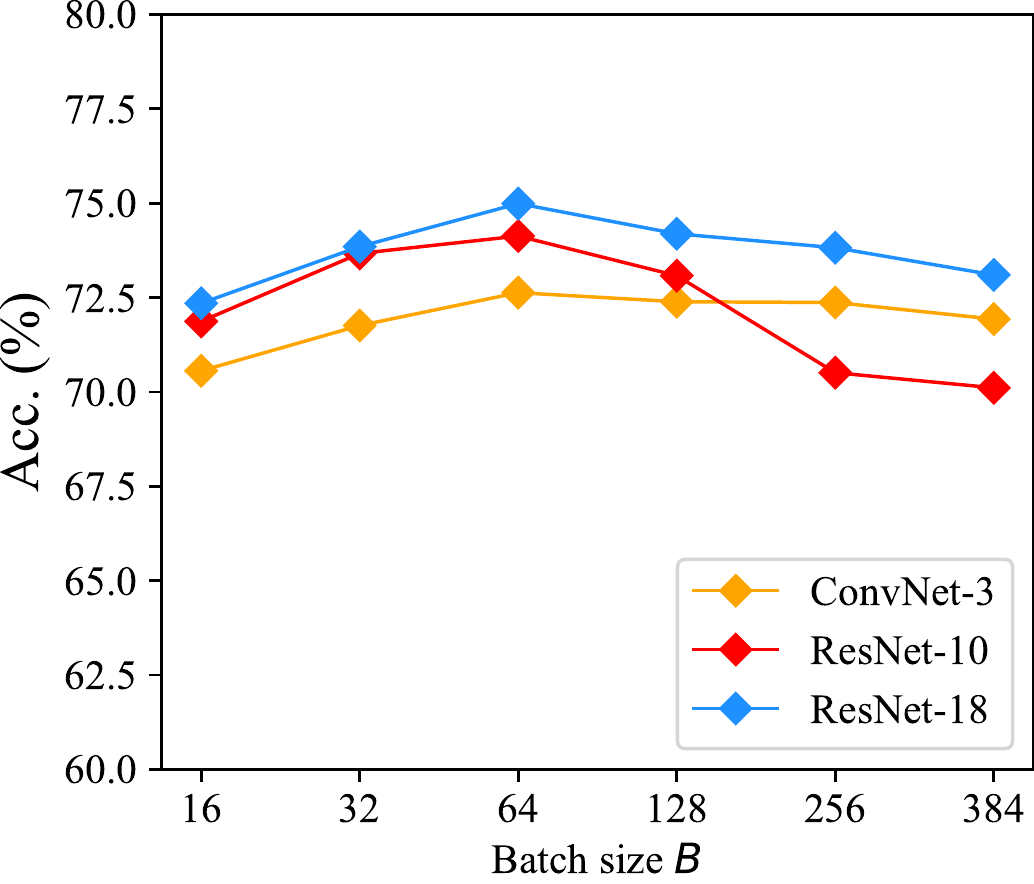}
    \caption{Ablation of batch size.}
    \label{fig:abl_bz}
    \end{subfigure}
    \begin{subfigure}{0.33\textwidth}
    \raisebox{0.0mm}{\includegraphics[width=0.95\linewidth,height=0.82\linewidth]{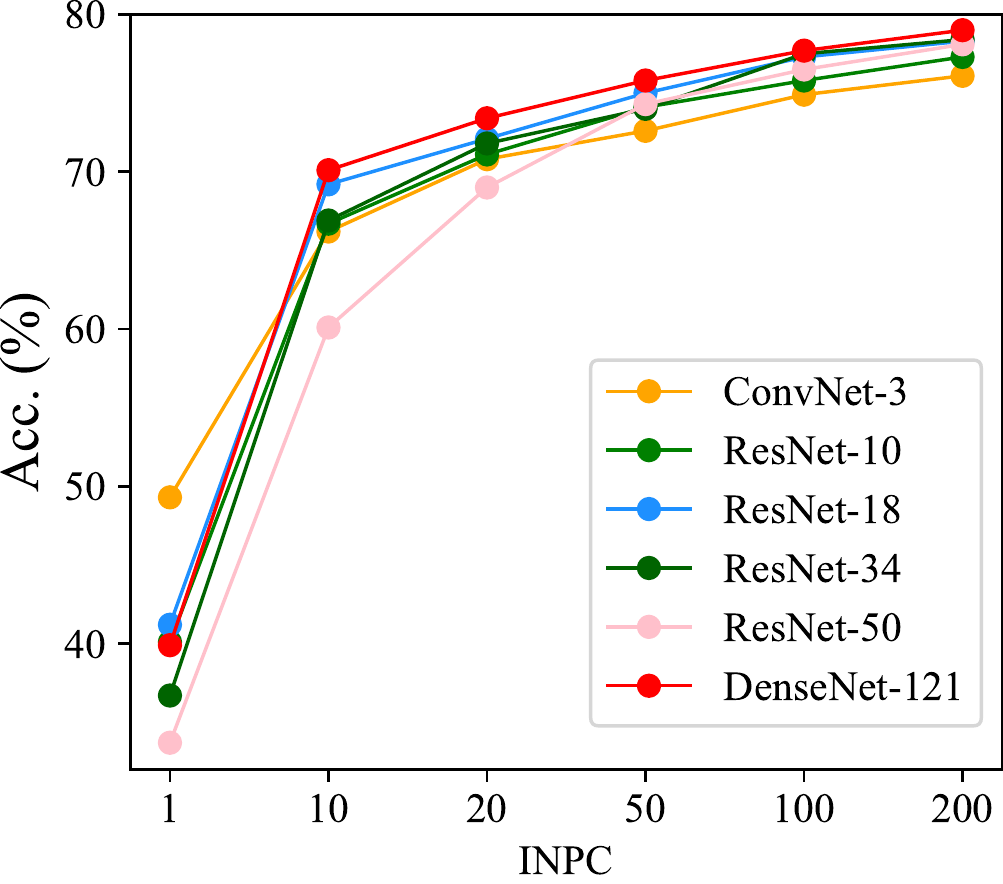}}
    \caption{Evaluation of scalability and efficiency.}
    \label{fig:efficiency_scalability}
\end{subfigure}
\caption{(a) evaluates the sensitiveness of $\lambda$ in DiM with ConvNet-3 (C-3) and ResNet-10 (R-10). (b) shows ablations of different batch sizes on C-3, R-10, and R-18. (c) explores scalability to large architectures and re-deployment efficiency of DiM on C-3, R-10, R-18, R-34, R-50, and DenseNet-121. All the results are implemented on CIFAR-10 with INPC = 50.
DiM achieves the highest results with $\lambda$ = 0.01 and $B$ = 64. Our method also peform strong scalability to large architectures and high efficiency of redeployment. Best viewed in color.}
\end{figure*}

\begin{table}[t]
    \centering
    \setlength{\tabcolsep}{7pt}
    \caption{Evaluation of feature matching, gradient matching, and logits matching strategies in the proposed framework. \textbf{Bold entries} are best results.}
    \begin{tabular}{ccc|ccc}
        \toprule
        \multicolumn{3}{c|}{Match Strategy} & \multicolumn{3}{c}{Acc.} \\
         feature & gradient & logits & C-3 & R-10  & R-18\\
         \midrule
         &&  & 68.1 & 68.7 &69.1\\
         \checkmark & & &72.0 & 73.2 &73.8 \\
         & \checkmark & & 71.3 & 71.1 &71.7\\
          & &\checkmark&\textbf{72.6} & \textbf{74.1} & \textbf{75.0} \\
         \bottomrule
    \end{tabular}
\label{tab:abl_matching_or_not}
\end{table}

\textbf{Evaluation of matching strategies.} Vanilla GANs synthesize real-looking images, but it lacks supervision for generating discriminative images.
Here, we investigate the effects of three matching strategies (feature, gradient, and logits matching) for DiM on CIFAR-10. As shown in Tab. \ref{tab:abl_matching_or_not}, one can find that three matching strategies consistently improve vanilla GANs with 3\% $\sim$ 6\% on ConvNet-3 (C-3), ResNet-10 (R-10), and ResNet-18 (R-18). It proves matching strategies enhance the discrimination of generated images.
Among them, gradient matching obtains the worst performances.
We attribute it to the fact that the gradient difference fails to provide proper supervision when the distribution gap between synthetic and real images is small. 
For example, even for two random batches from the original dataset, the gradient difference is still large. 
Feature matching only constrains the overall distribution of synthetic images, so it performs worse than our logits matching.
In DiM, the logits of original batches are regarded as soft labels for synthetic images, which provides more abundant supervision for the generator~\cite{hinton2015distilling}.
Meanwhile, logits matching is not required to be optimized category by category, which largely improves the training efficiency.  

\begin{table}[t]
    \centering
    \caption{Exploring which epoch(s) to perform logits matching. The best result is marked in \textbf{bold}.}
    \label{tab:when_to_perform_match}
    \begin{tabular}{c|c|c|c|c|c|c}
        \toprule
        Acc. / $\mathbf{N}$ & 20 & 50 & 80 & 100 & 120 & 150\\
        \midrule
        C-3 (\%) &65.6  &70.4  &71.7 &71.8 &\textbf{72.6} &72.5 \\
        R-10 (\%) &67.0  &71.4  &73.0 &73.2 &\textbf{74.1} &74.1 \\
        \bottomrule
    \end{tabular}
\end{table}
\textbf{Exploring when to perform matching.}
\label{sec:matching_time}
$N$ is a hyperparameter that determines in which epoch starts the training with logit matching loss. We vary $N$ from 20 to 150 and show the results in Tab. \ref{tab:when_to_perform_match}. We report the results of ConvNet-3 (C-3) and ResNet-10 (R-10), respectively. The performance goes up gradually from the 20th to the 120th epoch. The best result is achieved at $N = 120$. After the 120th epoch, the performances converge in the range around 72.5 on C-3 and 74.1 on R-10.
It can be explained by that adding logits matching too early disturbs the optimization of GANs and too late obtains few extra gains from GANs.

\begin{table}[t]
\caption{Explorations of the diversity of models pool. C-3 denotes ConvNet-3. We utilize both average and max pooling for ResNet-10 (R-10) and ResNet-18 (R-18). \textbf{Bold entries} are best results.}
\label{tab:models_pool}
    \centering
    \small
    \setlength{\tabcolsep}{8pt}
    \begin{tabular}{ccc|ccc|c}
        \toprule
        \multicolumn{3}{c|}{L-Match. Models} &\multicolumn{3}{c|}{ Deployment Acc. (\%)} &\multirow{2}{*}{Avg.}\\
         C-3 & R-10 & R-18 &  C-3 & R-10 & R-18 \\
         \midrule
         \checkmark & & &72.0 &73.5&73.9 &73.1   \\
         &\checkmark & &72.1 &73.2&74.2 &73.2 \\
         & &\checkmark &72.2 &73.4&74.3 &73.3 \\
          \checkmark &\checkmark & &72.4 &73.6&73.7  &73.2\\
          \checkmark & &\checkmark &72.4&73.4&74.5  &73.4 \\
         &\checkmark &\checkmark &72.5 &73.6&74.6  &73.6\\
         \checkmark&\checkmark &\checkmark &\textbf{72.6} &\textbf{74.1}&\textbf{75.0} &\textbf{73.9} \\
         \bottomrule
    \end{tabular}

\end{table}
\textbf{Exploring the diversity of models pool.}
We expect the model pool to provide various supervision that facilitates to synthesize images helpful for downstream classification tasks. To explore the effect of the diversity in the models pool, we design an experiment to ablate it in Tab. \ref{tab:models_pool}. 
We employ three architectures (C-3, R-10, R-18) in our models pool for logits matching.
For testing, we evaluate trained $\mathcal{G}$ on random initializated architectures.
The high diversity models pool obtains higher results than low diversity ones, which demonstrates various supervisions are crucial for the generator to distill information from the original dataset. Another interesting finding is DiM can distill datasets using a small-scale architecture and be deployed with large-scale architectures to obtain higher downstream performances, such as distillation with C-3 and deployment on R-10 or R-18.
It largely saves the cost of the most time-consuming re-distillation.
We also investigate our DiM with more complex architectures, such as R-10 and R-18. Previous results in~\cite{kimICML22, zhao2021DC, zhao2021dsa, wang2022cafe} show that matching with complex architectures perform worse than C-3 in DD settings. 
However, our DiM performs better matched with complex architectures, which is the first time the performance on complex architectures outperforms simple architectures.

\begin{table}
    \centering
    \caption{Illustration of the performance (\%) on unseen architectures. DiM shows strong generalization on various architectures. The distillation architecture is ConvNet-3. All results are evaluated on CIFAR-10 with 10 INPC or IPC. \textbf{Bold entries} are best results.}
    \label{tab:cross-arch}
    \resizebox{0.99\linewidth}{!}{
    \begin{tabular}{cc|cccc}
        \toprule& &\multicolumn{4}{c}{Evaluation Model}\\
        & & ConvNet-3 & ResNet-18 & VGG-11 & AlexNet  \\
        \midrule \multirow{4}{*}{\rotatebox[origin=c]{90}{Method}} &
        DiM & \textbf{66.2 $\pm$ 0.5} & \textbf{69.2 $\pm$ 0.3} & \textbf{66.8 $\pm$ 0.5} & \textbf{67.3 $\pm$ 0.9}\\
        &FRePo & 65.5 $\pm$ 0.4 & 57.7 $\pm$ 0.7 & 59.4 $\pm$ 0.7 & 61.9 $\pm$ 0.7\\
        &MTT & 64.3 $\pm$ 0.7 & 46.4 $\pm$ 0.6 & 50.3 $\pm$ 0.8 & 34.2 $\pm$ 2.6\\
        &DSA & 52.1 $\pm$ 0.4 & 42.8 $\pm$ 1.0 & 43.2 $\pm$ 0.5 & 35.9 $\pm$ 1.3\\
        &KIP & 47.6 $\pm$ 0.9 & 36.8 $\pm$ 1.0&  42.1 $\pm$ 0.4& 24.4 $\pm$ 3.9 \\\bottomrule
    \end{tabular}
    }
\end{table}
\textbf{Cross-Architecture Generalization.}
Previous works \cite{wang2022cafe,zhao2021DC, zhao2021dsa, nguyen2020dataset, cazenavette2022distillation} show the cross-architecture generalization among ConvNet-3~\cite{gidaris2018dynamic}, ResNet-18~\cite{he2016deep}, VGG-11~\cite{simonyan2014very}, and AlexNet~\cite{krizhevsky2017imagenet}. They distill on CIFAR-10 using ConvNet-3 and evaluate synthetic images on remained architectures. We compare cross-architecture generalization performances to previous state-of-the-art methods in Tab. \ref{tab:cross-arch}.
We follow the setting of MTT~\cite{cazenavette2022distillation} to evaluate cross-architecture generalization on CIFAR-10 with INPC=10.
DiM shows a significant improvement in cross-architecture generalization. Specifically, DiM outperforms MTT~\cite{cazenavette2022distillation}, DSA~\cite{zhao2021dsa}, FrePo~\cite{zhou2022dataset}, and KIP~\cite{nguyen2021dataset} with 6\% $\sim$ 43\% among all evaluation architectures. Thanks to the strong information capacity of the generative model, DiM is the first work where ResNet-18 performs better than ConvNet-3.

\textbf{Evaluation of $\lambda$.} 
\label{sec:lambda}
As default, logits matching loss $L_{m}$ is added to train generator from 120th epoch.
We evaluate the sensitiveness of the trade-off hyper-parameter $\lambda$ between $L_{g}$ and $L_{m}$. As shown in Fig. \ref{fig:sensitive_of_lbd}, one can find that $\lambda = 0.01$ perform the highest result on CIFAR-10. 
Too large $\lambda$, such as $\lambda = 1.0$, reduces the effect of $L_{g}$, so the semantic information of synthetic images can not be guaranteed. Conversely, too small $\lambda$ degrades the generator to vanilla GANs. Thus the discrimination of synthetic images may be lost. To better understand the relation between synthetic images and $\lambda$, we show more visualizations in the supplementary file. 

\textbf{Exploring the influence of $B$.} 
\label{sec:batch_size}
We use $B$ to represent the batch size of real images during the logits matching. The experiment results are shown in Fig. \ref{fig:abl_bz}. To explore the general characteristic of $B$, we study the sensitiveness of $B$ from 16 to 384 on C-3, R-10, and R-18. 
Our defaulted $\mathbf{B} = 64$ consistently obtains the highest result with C-3, R-10, and R-18. Too large $\mathbf{B}$ provides more information in each iteration, but it also increases the difficulties of optimization.
Therefore, the performance drops significantly.
Too small $B$, such as $B = 16$, can only provide limited supervision from the original dataset in each iteration. Thus it also performs poorly.

\begin{table}[t]
    \centering
    \small
    \setlength{\tabcolsep}{12pt}
    \caption{Performance comparison to GANs. }
    \label{tab:compare_to_gan}
    \begin{tabular}{c|c|c}
        \toprule
        INPC & C-3 / R-10 + GANs & C-3 / R-10 + DiM  \\
        \midrule
        1 &46.4 / 38.9  &\textbf{51.3 / 42.5}    \\
        10 &62.7 / 60.7  &\textbf{66.2 / 66.7}  \\
        50 &68.1 / 68.7  &\textbf{72.6 / 74.1}   \\
        \bottomrule
    \end{tabular}
\end{table}

\textbf{Evaluating the effort of generation in DiM.}
Compared to previous works~\cite{zhao2021DC, wang2022cafe, cazenavette2022distillation, kimICML22, yang2022dery}, DiM needs to generate training images on the fly. To evaluate the effort of this step, we design a experiment to compare the time cost of image generating and training on different architectures.
As shown in Tab. \ref{tab:effort}, the extra cost of DiM is light and can be ignored in most cases (R-50 and D-121).
The effort of images generation takes up 2\% $\sim$ 6\% of the whole cost on D-121, R-50, and R-34.
Even on R-18, R-10, and C-3, the cost of generator only takes up 11\% $\sim$ 20\%.

\begin{table}[t]
    \centering
    \small
    \caption{Evaluating the effort of generator. We show the effort of image generation and training separately. Gen. denotes the generation step. Performed with batch size of 128.}
    \label{tab:effort}
    \begin{tabular}{c|c|c|c|c|c|c}
        \toprule
        Part / Arch. & C-3 & R-10 & R-18 & R-34 & R-50 & D-121\\
        \midrule
        Gen. (ms) &2  &2  &2 &2 &2 &2 \\
        Train (ms) &8  &12  &17 &32 &45 &80 \\
        Ratio &0.20  &0.14  &0.11 &0.06 &0.04 &0.02 \\
        \bottomrule
    \end{tabular}
\end{table}

\textbf{Scalability to large model and efficiency of re-deployment.} As aforementioned, there are mainly two limitations of previous works. The first one is poor scalability. For example, the images synthesized by small architectures always perform worse on larger ones \cite{zhao2021DC, zhao2021dsa, wang2022cafe}.
The second one is low re-deployment efficiency. Previous DD methods \cite{zhao2021DC, zhao2021dsa, wang2022cafe} need to re-distill when INPC changes, which is time-consuming and results in high computational costs.
To investigate the architecture scalability and re-deployment efficiency of DiM, we evaluate DiM on ResNets~\cite{he2016deep} and DenseNet-121~\cite{huang2017densely} with INPC = 1, 10, 20, 50, 100, 200 in Fig. \ref{fig:efficiency_scalability}.
We are able to conclude that:
i). The model capacity matches the size of INPC. Simple architectures perform better with small INPC, while complex architectures perform better with large INPC.
For example, under INPC = 1, ConvNet-3 outperforms ResNet-34, ResNet-50, and DenseNet-121 by 12.6\%, 15.8\%, and 9.4\%, respectively. 
ii). Except INPC = 1, DiM shows strong scalability to larger model, while previous works drop largely when applied on large model, see in Tab. \ref{tab:cross-arch}.
iii). As the optimization is conducted on generative models instead of images, our Dim is only required to be trained once for various subsequent applications. (\textit{i.e}. no re-distillation requirement). Meanwhile, the performance on INPC=100, 150, and 200 increases significantly than that on small INPC.

\textbf{Comparisons to GANs. }
We compare the performances of 1, 10, and 50 INPC of GANs and DiM on ConvNet-3 (C-3) and ResNet-10 (R-10). As shown in Tab. \ref{tab:compare_to_gan}, DiM achieves the highest results in all cases. Compare to GANs, the improvement of DiM is up to 5\%, which indicates DiM distills more discriminative knowledge from original train dataset. This conclusion is also aligned with our analysis: GAN aims to generate images that look real while DiM targets for synthesizing samples helpful for downstream classification training.

\section{Conclusion and Discussion}
In this work, we propose a novel paradigm to distill information from original train set to model instead of synthetic images, termed as DiM.
The DiM consists of a models pool to provide various supervisions for distilling, and a logits matching loss that minimizes the logits differences between real and synthetic images.
The simple yet effective design brings DiM significant improvements over previous DiI methods in large architecture scalability, redeployment efficiency, and cross-architecture generality.

\textbf{Limitations and Broader Impacts.}
Our DiM needs to generate training samples at the deployment stage, which leads to extra efforts for downstream tasks. We also evaluate the influence of it in supplementary materials. In the future, we aim to design light-weighted generative model for DiM. Meanwhile, we plan to explore the effects of DiM on large-scale dataset such as ImageNet \cite{deng2009imagenet} and other tasks such as object detection ~\cite{lin2014microsoft}, semantic segmentation ~\cite{zhou2017scene}, video understanding~\cite{soomro2012ucf101}, and so on. The proposed method does not contain any studies involving affecting ethics or human rights performed by any of the authors. 

{\small
\bibliographystyle{ieee_fullname}
\bibliography{egbib}
}

\newpage
\section{More Evaluations and Comparisons}
In this section, we provide more evaluations and comparisons of DiM.
Firstly, we explore the influence of the dimension of random noise $K$. 
Secondly, we evaluate the effort of image generation in deployment stage.
Finally, the details of implementation hardware are shown.

\textbf{Evaluation of $K$.} 
We default to set the dimension of $K$ = 100 for the conditional GAN~\cite{mirza2014conditional} in experiments. To investigate the influence of the dimension, we design an ablation of $K$. As shown in Tab. \ref{tab:abl_k}, the defaulted $K$ = 100 achieves the highest results on ConvNet-3 (C-3), ResNet-10 (R-10), and ResNet-18 (R-18).
For $K$ from 32 to 100, the average performance increases from 71.8\% to 73.9\% and drops gradually when $K > 100$. Both too small and too large $K$ leads difficulties for optimization.

\begin{table}[h]
    \centering
    \caption{Exploring the dimension of $K$. The best result is marked in \textbf{bold}.}
    \label{tab:abl_k}
    \begin{tabular}{c|c|c|c|c|c}
        \toprule
        Acc. / Dim. of $K$ & 32 & 64 & 100 & 128 & 200 \\
        \midrule
        C-3 (\%) &70.5  &71.0   &\textbf{72.6} &68.9 &68.2  \\
        R-10 (\%) &72.3  &72.5  &\textbf{74.1} &69.7 &69.4 \\
        R-18 (\%) &72.6  &72.8  &\textbf{75.0} &72.1 &70.2 \\
        \midrule
        Avg. Acc. &71.8 &72.1 &\textbf{73.9}&70.2&69.3\\
        \bottomrule
    \end{tabular}
\end{table}



\textbf{Details of implementation hardware}
We perform all experiments on single Tesla V100 32G GPU of Linux Cluster with a 2.40 GHz
Intel (R) Xeon (R) Platinum 8260 CPU (24 cores).

\section{More Visualizations}

\begin{figure*}[htp]
    \centering
    \includegraphics[width =\textwidth]{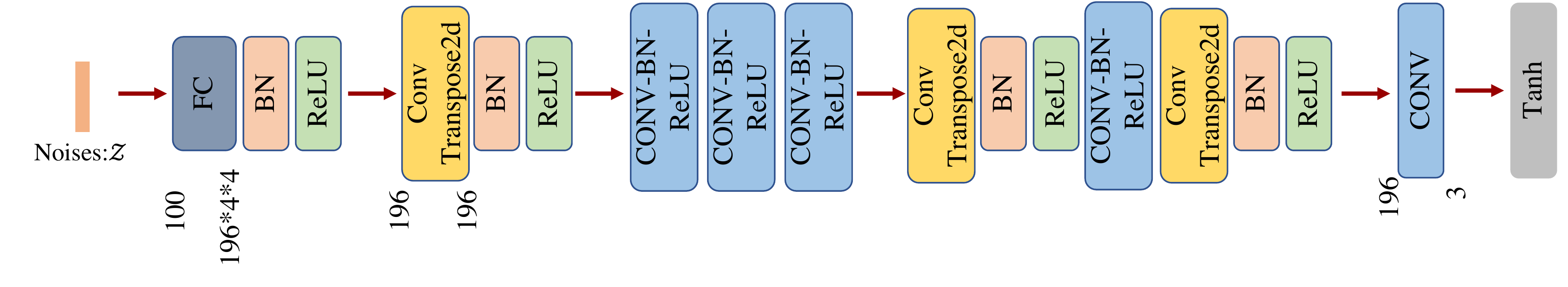}
    \vspace{-2em}
    \caption{Visualization of the architecture of generator in DiM.}
    \label{fig:gan_arch}
\end{figure*}

\textbf{The architecture of GAN.}
To better understand our DiM, we also visualize the detailed architecture of the generator.
As illustrated in Fig. \ref{fig:gan_arch}, the generator consists of a Fully-Connected (FC) layer, BN, ReLU, 3 ConvTranspose2d layers, BN, ReLu, 4 blocks of CONV-BN-ReLU, CONV, and tanh activation.
The input and output dimensions of FC are 100 and 194*4*4. The input and output channels of the last CONV are 196 and 3. More details can be checked from the zip file of our code.

\begin{figure*}[htp]
    \centering
    \includegraphics[width =\textwidth]{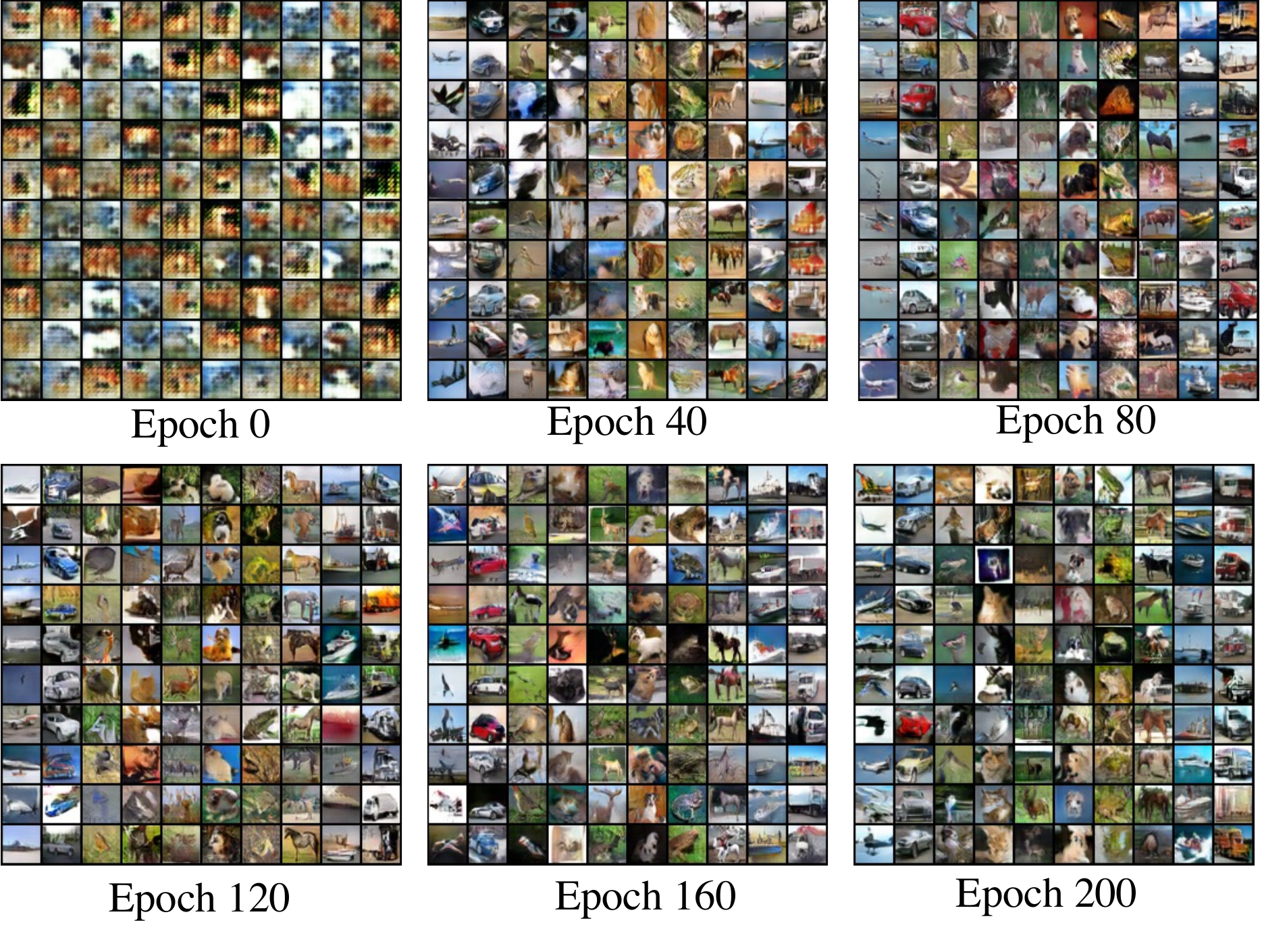}
    \caption{Visualization of synthetic images during training. We show the 10 images per class synthetic set as an example.}
    \label{fig:training_process}
\end{figure*}

\textbf{Synthetic images during the training.}
We visualize the synthetic images during the training process and show them in Fig. \ref{fig:training_process}.
One can find that the image quality gradually increase with the training.

\begin{figure*}[htp]
    \centering
    \includegraphics[width =\textwidth]{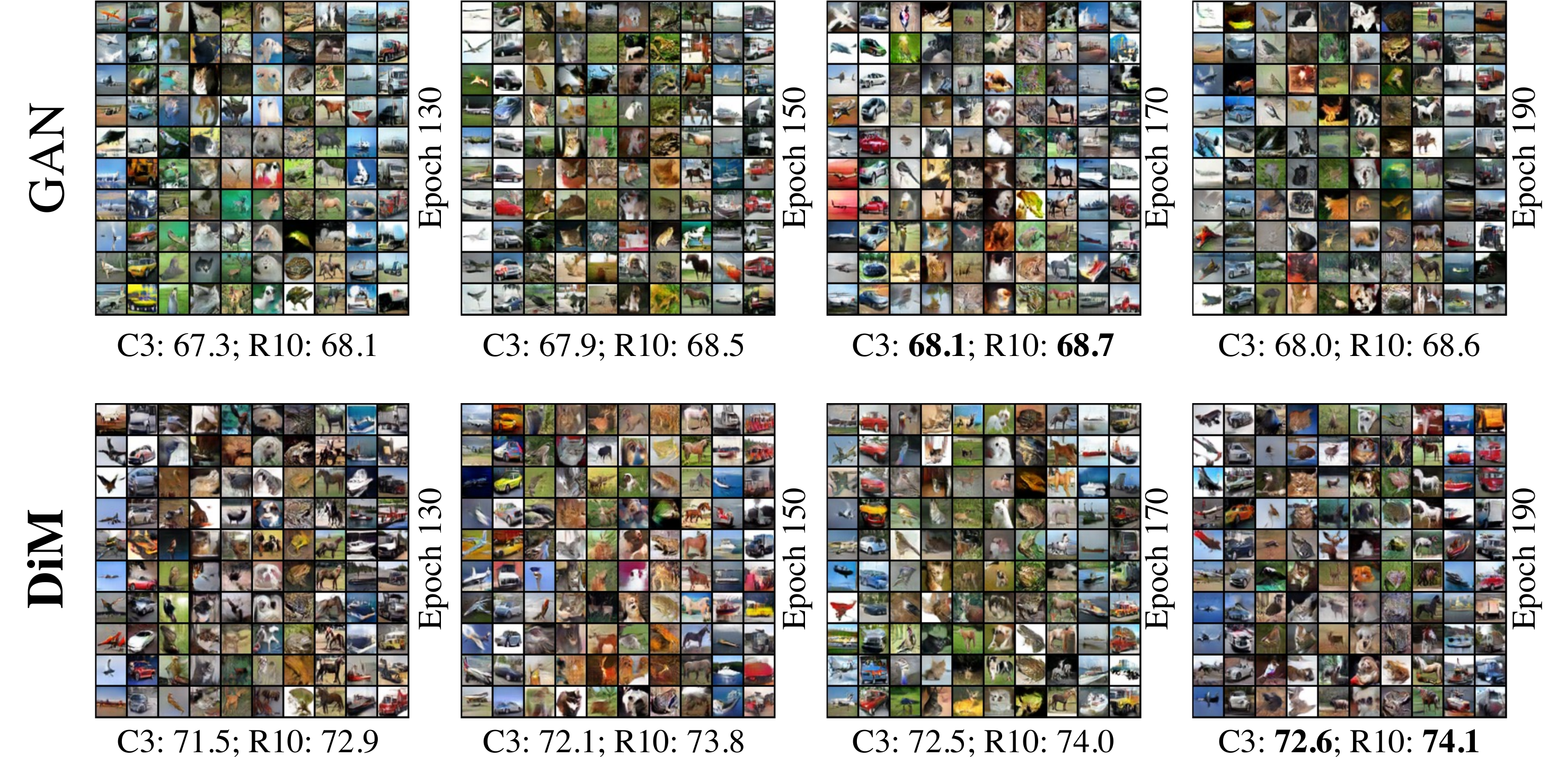}
    \caption{Visualization of synthetic images during training. Due to the limitation of space, we only show the 10 images per class synthetic set as an example.}
    \label{fig:gan_vs_dim}
\end{figure*}

\textbf{Synthetic images of GANs and DiM.} 
We visualize the synthetic images during the training and the evaluation performances of GAN and DiM, respectively.
We show the results of C-3 and R-10 at 130, 150, 170, and 190 epochs.
As shown in Fig. \ref{fig:gan_vs_dim}, DiM outperforms GAN at all epochs with 4\%$\sim$5.5\%, which indicates the strong discrimination of training images generated by DiM. 

\begin{figure*}[htp]
    \centering
    \includegraphics[width =\textwidth]{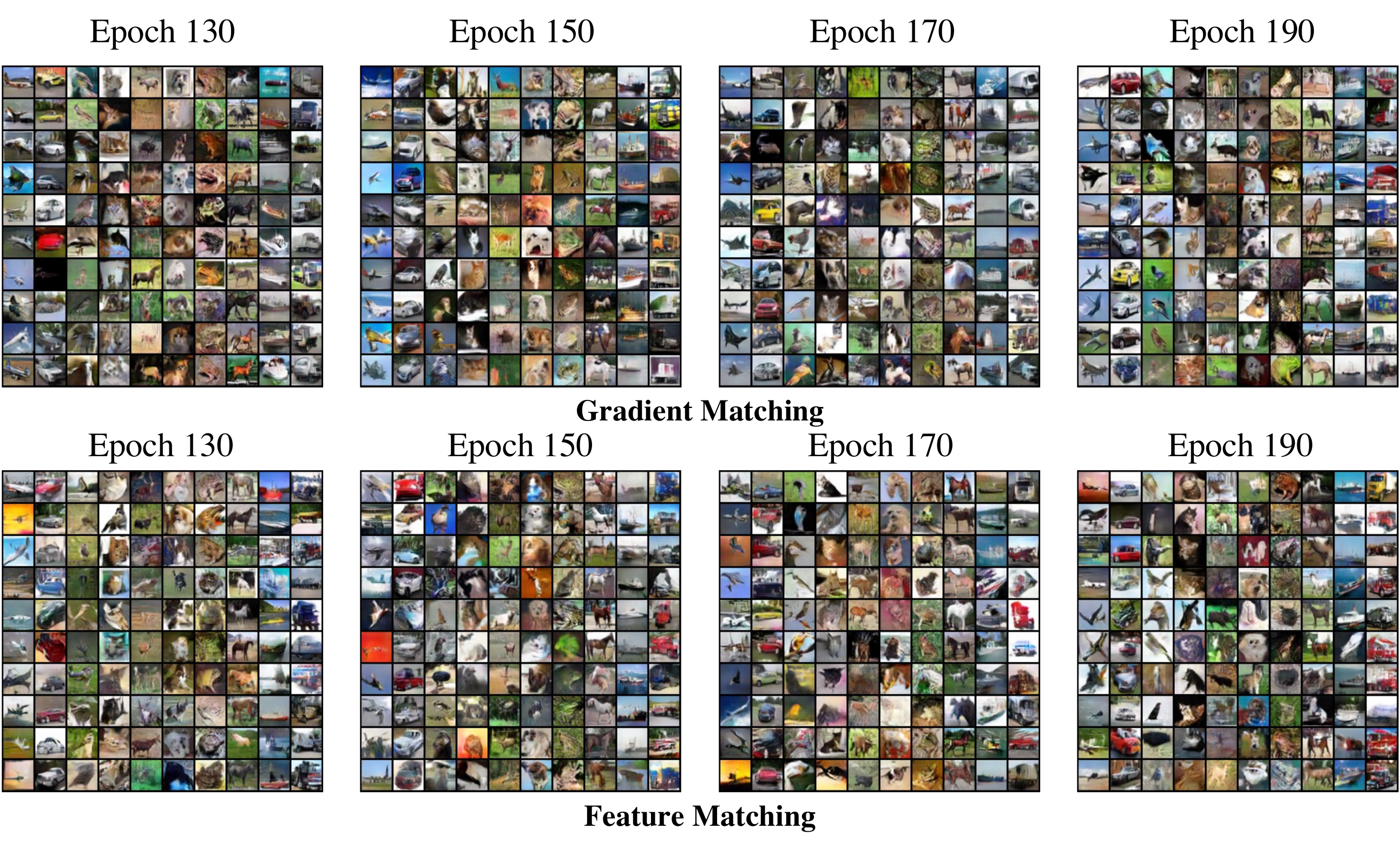}
    \caption{Visualization of the images generated by gradient and feature matching at 130, 150, 170, and 190 epochs.}
    \label{fig:matching_comparison}
\end{figure*}

\textbf{Generated images with other matching strategies.}
Previous works~\cite{zhao2021DC, zhao2021dsa, wang2022cafe, kimICML22} have explored gradient and feature/distribution matching strategies. We also compare our logits matching to these strategies in experiment section. The images generated by gradient and feature matching are shown in Fig. \ref{fig:matching_comparison}.

\end{document}